\documentclass[11pt]{article}
\usepackage{eacl2017}
\usepackage{amssymb}
\usepackage{graphicx}
\usepackage{times}
\usepackage{url}
\usepackage{latexsym}
\usepackage{amsmath}
\usepackage{amssymb}
\usepackage{graphicx}
\usepackage{subfig}
\usepackage{framed,multirow}
\usepackage{algorithm2e}
\usepackage{float}
\usepackage{wrapfig}
\usepackage{booktabs}
\restylefloat{figure}
\usepackage[capitalize]{cleveref} 

\eaclfinalcopy 

\graphicspath{{Grafica/}}

\title{Neural Networks Classifier for Data Selection\\in Statistical Machine Translation}

\author {\'{A}lvaro Peris, Mara Chinea-Rios, Francisco Casacuberta \\
  PRHLT Research Center \\Universitat Polit\`ecnica de Val\`encia, Valencia (Spain)\\
    {\tt \{lvapeab,machirio,fcn\}@prhlt.upv.es}}

\date{}

\begin{document}
\maketitle

\begin{abstract}
We address the data selection problem in statistical machine translation (SMT) as a classification task. The new data selection method is based on a neural network classifier. 
We present a new method description and empirical results proving that our data selection method provides better translation quality, compared to a state-of-the-art method (i.e., Cross entropy). Moreover, the empirical results reported are coherent across different language pairs. 
\end{abstract}

\section{Introduction}\label{sec:introduction}
The performance of a SMT system is dependent on the quantity and quality
of the available training data. Typically, SMT systems are trained with all available data, assuming that the more data used to train the system, the better. 
This assumption is backed by the evidence that scaling to ever larger data shows continued improvements in quality, even when one trains models over billions of n-grams~\cite{Brants2007}. In the SMT context, n-grams refers to sequences of $n$ consecutive words.
However, growing the amount of data available is only feasible to a certain extent. 
Moreover so, whenever it is critical that such data is related to the task at hand.
In fact, translation quality is negatively affected when there is insufficient training data for the specific domain to be tackled in production conditions \cite{Callison2007meta,koehn2010statistical}.

Data selection (DS) has the aim of selecting for training, the best subset of sentence pairs from an available pool, so that the translation quality achieved in the target domain is improved. 

State-of-the-art DS approaches rely on the idea of choosing those sentence pairs in an out-of-domain training corpus that are in some way similar to an in-domain training corpus in terms of some different metrics. Cross-entropy difference is a typical ranking function \cite{Moore2010,Axelrod2011,Xenc,Schwenk2012,Mansour2011}.

On the other hand, distributed representations of words have proliferated during the last years in the research community. Neural networks provide powerful tools for processing text, achieving success in text classification \cite{Kim14} or in machine translation \cite{Sutskever14,Bahdanau15}. Furthermore, \cite{Duh2013} leveraged neural language models to perform DS, reporting substantial gains over conventional n-gram language model-based DS.

Recently, convolutional neural networks (CNN) have also been used in data selection \cite{Chen16,Chen16b}. In these works, the authors used a similar strategy to the one proposed in Section~\ref{sec:sss}, but in a different scenario: they have no in-domain training corpus; only a large out-of-domain pool and small sets of translation instances. Their goal was to select, from the out-of-domain corpus, the more suitable samples for translating their in-domain corpora.

This paper tackles DS by taking advantage of neural network as sentence classifiers with the ultimate goal of obtaining corpus subsets that minimize the training corpus size, while improving translation quality.

The paper is structured as follows. Section~\ref{sec:DS-method}, presents our DS method, featuring two different neural network architectures: a CNN \cite{LeCun98} and a bidirectional long short-term memory (BLSTM) network \cite{Hochreiter97,Schuster97}. In Section~\ref{sec:experiments}, the experimental design and results are detailed. Finally, the main results of the work and future work are discussed in Section~\ref{sec:conclusions}.

\section{Data Selection}\label{sec:DS-method}
The goal of DS methods consists in to select a subset $S$ of source language sentences from an out-of-domain pool of sentences $G$ based on a source in-domain corpus $I$.

\subsection{Data Selection using Cross-entropy}
\label{sec:CE_method}
As mentioned in Section \ref{sec:introduction}, one well-established DS method consists in scoring the sentences in the out-of-domain corpus by their perplexity. Moore and Lewis \shortcite{Moore2010} used cross-entropy rather than perplexity, even though they are both monotonically related. 
For selecting $S$, this technique relates the cross-entropy given by a language model trained over the $I$ together with the out-of-domain ($G$) language model, computing a score for a sentence $\mathbf{x}$: 
\begin{equation}
\label{eq:cross-entropy}
c(\mathbf{x}) = H_{I} (\mathbf{x}) - H_{G}(\mathbf{x})
\end{equation}
where $H_{I}$ and $H_{G}$ are the in-domain and out-of-domain cross-entropy of sentence $\mathbf{x}$, respectively.
\subsection{Data Selection using Neural Networks} 

In this work, we tackle the DS problem 
as a classification task. Let us consider a classifier model $M$ that assigns a probability $p_M (I|\mathbf{x})$ to a sentence $\mathbf{x}$, depending whether $\mathbf{x}$ belongs to the in-domain corpus $I$ or not.

In this case, to obtain the selection $S$, one could just apply $M$ to each sentence from the out-of-domain pool $G$ and select the most probable ones.

\begin{wrapfigure}{R}{0.2\textwidth}
	\centering
	\includegraphics[width=0.2\textwidth]{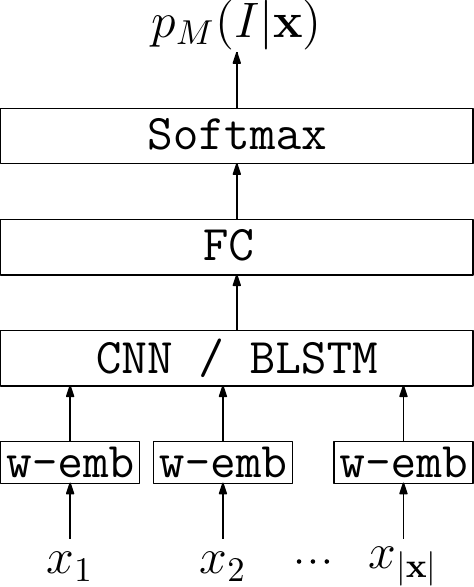}
	\caption{\label{fig:classifier} General architecture of the proposed classifier. {\tt w-emb} stands for word-embedding and {\tt FC} for fully-connected layer.}
\end{wrapfigure}  

We explored the use of CNN and BLSTM networks as sentence encoders. As shown in Fig.~\ref{fig:classifier}, the input sentence is fed to our system following a one-hot codification scheme and is projected to a continuous space by means of a word-embedding matrix. Next, the input sentence is processed either by a CNN or a BLSTM network. After this, we add one or more fully-connected (FC) layers. Finally, we apply a softmax function, in order to obtain normalized probabilities. All elements from the neural classifier can be jointly trained by maximum likelihood.

\paragraph{Convolutional Neural Networks.}

CNNs have proven their representation capacity, not only in computer vision tasks \cite{Szegedy15}, but also representing text \cite{Kalchbrenner13,Kim14}. In this work, we used the non-static CNN proposed by Kim \shortcite{Kim14}.

This CNN consists in the application of a set of filters to windows of different length. These filters apply a non-linear function (e.g. ReLU). Next, a max-pooling operation is applied to the set of convolutional filters.  As result, the CNN obtains a feature vector representing the input sentence.

\paragraph{Recurrent Neural Networks.}

In recurrent neural networks, connections form a directed cycle. This allows the network to maintain an internal state and be powerful sequence modelers. Moreover, bidirectional networks have two independent recurrent layers, one processing the input sequence in a forward manner and other processing it a backward manner. Therefore, they allow to exploit the full context at each time-step. Gated units, such as LSTM~\cite{Gers00}, mitigate the vanishing gradient problem and hence, they are able tp properly model long sequences. 

BLSTM networks can be used for effectively encoding a sentence by concatenating the last hidden state of the forward and backward LSTM layers. This provides a compact representation of the sentence, which takes into account relationships in both time directions.

\subsection{Semi-supervised Selection}
\label{sec:sss}
Properly training these neural classifiers may be a challenging task, since the in-domain data is scarce. Hence, for training them, we follow a semi-supervised iterative protocol \cite{Yarowsky95}.

Since the data selection is a binary classification problem, we need a set of positive and negative training samples. We start from an initial set of positive samples $P_0$ and a set of negative samples $N_0$ and train an initial model $M_0$. At each iteration $i \geq 0$, we classify all sentences belonging to the out-of-domain pool (${G}_i$). We extract a number $r$ of top-scoring sentences and include them into the set of positive samples, producing a new set of positive samples $P_{i+1}$. Analogously, the $r$ bottom-scoring sentences are included into a new negative samples set $N_{i+1}$. Hence, at each iteration, we remove $2r$ samples from the out-of-domain set, producing the pool ${G}_{i+1}$. Then, we start a new iteration,  
training a new model $M_{i+1}$ with $\{P_{i+1} \cup N_{i+1}\}$. This is repeated until there are no more sentences in the out-of-domain pool. 

We set our in-domain corpus $I$ as $P_0$. We randomly extract $|I|$ sentences from $G$ for constructing $N_0$. The initial out-of-domain pool $G_0$ is defined as $\{{G} - N_0\}$.

\section{Experiments in SMT} \label{sec:experiments}
In this section, we empirically evaluate the DS strategy proposed in Section~\ref{sec:DS-method}. 
We conducted experiments with different language pairs with the purpose of evaluating whether the conclusions drawn from one single language pair hold in further scenarios.

\subsection{Corpora}
Two corpora are involved within the DS task: an out-of-domain corpus $G$ and an in-domain corpus $I$. DS selects only a portion of the out-of-domain corpus, and leverages that subset together with the in-domain data to train a, hopefully improved, SMT system. 
As out-of-domain corpus, we used the Europarl 
corpus \cite{koehn2005europarl}. As in-domain data, we used the EMEA corpus
~\cite{corpus-emea}. 
The Medical-Test and Medical-Mert corpora are partitions established in the 2014 Workshop on SMT\footnote{\tt www.statmt.org/wmt14/} \cite{WMT2014}.
We focused on the English-French (En-Fr), French-English (Fr-En), German-English (De-En) and English-German (En-De) language pairs.  The main figures of the corpora used are shown in Table~\ref{tab:corpora}.

\begin{table} [h]
\caption{\label{tab:corpora} Corpora main figures. (EMEA-Domain) is the in-domain corpus, (Medical-Test) is the evaluation data and (Medical-Mert) is the development set. (Europarl) is the out-of-domain corpus. M denotes millions of elements and k thousands of elements, $\vert S \vert$  stands for number of sentences, $\vert W \vert$  for number of words and $\vert V \vert$ for vocabulary size.}
\centering
\small
\begin{tabular}{lllll}
\toprule
\multicolumn{2}{l}{Corpus} & $\vert S \vert$ & $\vert W \vert$ & $\vert V \vert$\\
\midrule
\multirow{2}{*}{EMEA-Domain} & EN & \multirow{2}{*}{1.0M} & 12.1M & 98.1k \\
 & FR &  & 14.1M & 112k \\
\midrule
\multirow{2}{*}{Medical-Test} & EN & \multirow{2}{*}{1000} & 21.4k & 1.8k \\
 & FR &  & 26.9k & 1.9k  \\
\midrule
\multirow{2}{*}{Medical-Mert} & EN & \multirow{2}{*}{501} & 9.9k & 979 \\
 & FR &  & 11.6k & 1.0k \\ 
\midrule
\multirow{2}{*}{Medical-Domain} & DE &
 \multirow{2}{*}{1.1M} & 10.9M & 141k \\
 & EN &  & 12.9M & 98.8k \\
\midrule
\multirow{2}{*}{Medical-Test} & DE & \multirow{2}{*}{1000} & 18.2k & 1.7k \\
 & EN &  & 19.2k & 1.9k \\ 
\midrule
 \multirow{2}{*}{Medical-Mert} & DE & \multirow{2}{*}{500} & 8.6k & 874 \\
 & EN &  & 9.2k & 979 \\ 
\midrule
\multirow{2}{*}{Europarl} & EN & \multirow{2}{*}{2.0M} & 50.2M & 157k \\
 & FR &  & 52.5M & 215k  \\
\midrule
\multirow{2}{*}{Europarl} & DE & \multirow{2}{*} {1.9M} & 44.6M & 290k \\
 & EN &  & 47.8M & 153k \\
\bottomrule
\end{tabular}
\end{table}


\subsection{Experimental Setup}

All neural models were initialized using the word-embedding matrices from word2vec\footnote{\url{code.google.com/archive/p/word2vec}}. These matrices were fine-tuned during training. Following Kim~\shortcite{Kim14}, we used filter windows of lengths 3, 4, 5 with 100 features maps each for the CNN classifier. For training it, we used Adadelta~\cite{Zeiler12} with its default parameters. The BLSTM network was trained with Adam~\cite{Ba14}, with a learning rate of $10^{-4}$. In order to have a similar number of parameters than in the CNN (48 million approximately), we used 300 units in each LSTM layer. 2 fully-connected layers of size 200 and 100 were introduced after the CNN and BLSTM (see Section~\ref{fig:classifier}). All neural models\footnote{Source code available at~\url{https://github.com/lvapeab/sentence-selectioNN}.} were implemented using Theano~\cite{Theano16}. The number of sentences selected at each iteration ($r$) should be empirically determined. In our experiments, we set $r=50,000$.

\begin{table*}[ht!]
	\caption{Summary of best results obtained. Columns denote, from left to right: selection strategy, BLEU, number of sentences, given in terms of the in-domain corpus size, and $(+)$ selected sentences.}\label{tab:best-result} 
\centering
\small
		\begin{tabular}{p{3cm}p{2cm}p{2cm}p{2cm}p{2cm}}
			\toprule
			& \multicolumn{2}{c}{EN-FR} & \multicolumn{2}{c}{FR-EN} \\
			
			Strategy & {\small BLEU} & \#\,Sentences &  {\small BLEU} & \#\,Sentences \\ \midrule
			\texttt{baseline-emea} & $28.6\pm0.2$ & $1.0$M & $29.9\pm0.2$ & $1.0$M \\ 
			\texttt{bsln-emea-euro} & $29.4\pm0.1$ & $1.0$M$+1.5$M & $32.4\pm0.1$  & $1.0$M$+1.5$M \\
			Random & $29.4\pm0.4$ & $1.0$M$+500$k & $32.3\pm0.3$ & $1.0$M$+500$k  \\ \midrule
			Cross-Entropy &  $29.8\pm0.1$ & $1.0$M$+450$k & $31.8\pm0.1$ & $1.0$M+$600$k  \\ 
			BLSTM & $\mathbf{29.9\pm0.3}$ & $1.0$M$+\mathbf{300}$k & $32.3\pm0.1$ & $1.0$M$+500$k \\ 
			CNN & $29.8\pm0.2$ & $1.0$M$+450$k & $ 32.3\pm0.2$ & $1.0$M$+\mathbf{350}$k \\ 
			\midrule

			& \multicolumn{2}{c}{DE-EN} & \multicolumn{2}{c}{EN-DE} \\ 
			Strategy & {\small BLEU} & \#\,Sentences &  {\small BLEU} & 
			\#\,Sentences \\ \midrule
			\texttt{baseline-emea} & $23.7\pm0.2$ & $1.0$M & $15.6\pm0.1$ & $1.0$M \\ 
			\texttt{bsln-emea-euro} & $26.2\pm0.3$ & $1.0$M$+1.5$M & $16.6\pm0.2$  & $1.0$M$+1.5$M \\
			Random & $25.5\pm0.1$ & $1.0$M$+600$k & $16.8\pm0.1$ & $1.0$M$+550$k  \\ 			\midrule
			Cross-Entropy &  $25.5\pm0.3$ & $1.0$M$+600$k & $16.8\pm0.2$ & $1.0$M+$500$k  \\ 
			BLSTM & $25.9\pm0.1$ & $1.0$M$+500$k & $\mathbf{17.1\pm0.2}$ & $1.0$M$+\mathbf{400}$k \\
			CNN  & $ 25.9\pm0.1$ & $1.0$M$+\mathbf{400}$k & $16.9\pm0.1$ & $1.0$M$+350$k \\ 
			\bottomrule
		\end{tabular}
\end{table*}

All SMT experiments were carried out using the open-source phrase-based SMT toolkit Moses \cite{Koehn2007}. 
The language model used was a 5-gram, standard in SMT research, with modified Kneser-Ney smoothing \cite{Kneser1995improved}, built with the SRILM toolkit \cite{Srilm2002}.
The phrase table was generated by means of symmetrised word alignments obtained with GIZA++ \cite{OchNey2003}. 
The log-lineal combination weights $\lambda$ 
were optimized using MERT (minimum error rate training)~\cite{Och2003}. Since MERT requires a random initialisation of $\lambda$ that often leads to different local optima 
being reached, every result of this paper constitutes  the  average  of  10  repetitions, with the purpose of providing robustness to the results. In the tables, $95\%$ confidence intervals of these repetitions are shown. SMT output was evaluated by means of BLEU \cite{papineni2002bleu}. 

We compared the selection methods with two baseline systems. The first one was obtained by training the SMT system with in-domain training data (EMEA-Domain data). We will refer to this setup with the name of \texttt{baseline-emea}. A second baseline experiment has been carried out with the concatenation of the Europarl corpus and EMEA training data (i.e., all the data available). We will refer to this setup as \texttt{bsln-emea-euro}. In addition, we also included results for a purely random sentence selection without replacement. 

\subsection{Experimental Results}

Table~\ref{tab:best-result} shows the best results obtained with our data selection using the two neural network architectures proposed (CNN and BLSTM) and cross-entropy method for each language pair (EN-FR, DE-EN, EN-DE). 

In EN-FR and EN-DE, FR-EN, translation quality using DS improves over {\tt bsln-emea-euro}, 
but using significantly less data ($20\%$, $23\%$ and $26\%$ of the total amount of out-of-domain data, respectively). In
the case of DE-EN, translation quality results are similar, but using only a $23\%$ of the data.
According to these results, we can state that our DS strategy is able to deliver similar quality than
using all the data, but only with a rough quarter of the data.

All proposed DS methods are mostly able to improve over random selection but in some cases differences are not significant. It should be noted that beating random is very hard, since all DS methods, including random, will eventually converge to the same point: adding all the data available. The key difference is the amount of data needed for achieving the same translation quality.

 Results obtained in terms of BLEU with our DS method are slightly better than the ones obtained with cross-entropy. However, cross-entropy requires significantly more sentences to reach comparable translation quality.
 
Lastly, CNN and BLSTM networks seem to perform similarly. Therefore, we conclude that both architectures are good options for this task. 


\section{Conclusion and Future Work} \label{sec:conclusions}

We developed a DS method, based on sentence classification techniques. The classifier is based on CNNs or BLSTM neural networks. We thoroughly evaluated it over four language pairs. Our method yielded better translation performance than the cross-entropy DS technique, requiring a minor amount of data. Additionally, we found that both CNN and BLSTM networks performed similarly, thus being both suitable sentence encoders.

As future work, we aim to delve into the usage of semi-supervised training strategies for the classifier. Ladder networks~\cite{Rasmus15} seem a promising tool. We should investigate how to include them in our pipeline. We should also investigate the application of one-shot learning strategies to a similar scenario, where only the text to translate is available. 
\bibliography{eacl2017}
\bibliographystyle{eacl2017}






\end{document}